%% file: iclr2026_conference.tex
\documentclass{article} 

\usepackage{subfiles}
\usepackage{booktabs}  
\usepackage{fixltx2e} 
\usepackage{graphicx}      
\usepackage{adjustbox}     
\usepackage[table]{xcolor}
\usepackage{caption} 
\usepackage{subcaption}  
\usepackage{multirow}   
\usepackage{algorithm}
\usepackage{algpseudocode}

\usepackage{amssymb} 
\usepackage{tikz}  

\usepackage{iclr2026_conference,times}

\input{math_commands.tex}

\usepackage{hyperref}
\usepackage{url}

\usepackage{authblk}

\title{Beyond Monolithic Rewards: A Hybrid and Multi-Aspect Reward Optimization for MLLM Alignment} 

\author{Radha Gulhane}
\author{Sathish Reddy Indurthi}
\affil{%
  Zoom Communications, Inc. \\
  \texttt{\{radha.gulhane, sathishreddy.indurthi\}@zoom.us}
}

\iclrfinalcopy 
\begin{document}

\maketitle

\begin{abstract}

Aligning multimodal large language models (MLLMs) with human preferences often relies on single-signal, model-based reward methods. Such monolithic rewards often lack confidence calibration across domain-specific tasks, fail to capture diverse aspects of human preferences, and require extensive data annotation and reward model training. In this work, we propose a hybrid reward modeling framework that integrates complementary reward paradigms: (i) model-based rewards, where a learned reward model predicts scalar or vector scores from synthetic and human feedback, and (ii) rule-based rewards, where domain-specific heuristics provide explicit correctness signals with confidence. Beyond accuracy, we further incorporate multi-aspect rewards to enforce instruction adherence and introduce a generalized length-penalty reward to stabilize training and improve performance. The proposed framework provides a flexible and effective approach to aligning MLLMs through reinforcement learning policy optimization. Our experiments show consistent improvements across different multimodal benchmarks when applying hybrid and multi-aspect reward modeling. Our best performing model in the 3B family achieves an overall average improvement of ~9.5\% across general and math reasoning tasks. Focusing specifically on mathematical benchmarks, the model achieves a significant average improvement of ~16\%, highlighting its effectiveness in mathematical reasoning and problem solving.

\end{abstract}

\section{Introduction}

\subfile{sections/introduction}

\section{Related Work}
\subfile{sections/related_work}

\section{Methodology: \textbf{H}ybrid and Multi-\textbf{A}spect \textbf{R}eward \textbf{M}odeling \textbf{O}ptimization (HARMO)} \label{sec:methodology}
\subfile{sections/reward_modeling}

\section{Experiment}
\subfile{sections/experiment_v1}

\section{Conclusion}
\subfile{sections/conclusion_v1}

\bibliography{iclr2026_conference}
\bibliographystyle{iclr2026_conference}

\appendix
\subfile{sections/appendix}

\end{document}

%% file: math_commands.tex
\usepackage{amsmath,amsfonts,bm}

\def\eqref#1{equation~\ref{#1}}

\def\1{\bm{1}}

\DeclareMathAlphabet{\mathsfit}{\encodingdefault}{\sfdefault}{m}{sl}
\SetMathAlphabet{\mathsfit}{bold}{\encodingdefault}{\sfdefault}{bx}{n}

\newcommand{\KL}{D_{\mathrm{KL}}}



%% file: sections/introduction.tex
The advent of Multimodal Large Language Models (MLLMs) has pushed the boundaries of artificial intelligence, enabling models to reason over and generate content that integrates text, images, and other modalities \citep{openai2024gpt4technicalreport, liu2023visualinstructiontuning}. A prevailing paradigm for aligning these powerful models with human preferences is Reinforcement Learning from Human Feedback (RLHF) \citep{christiano2017deep, ouyang2022training}. Typically implemented with Proximal Policy Optimization (PPO) \citep{schulman2017ppo}, RLHF fine-tunes a model's policy by optimizing a signal from a learned Reward Model (RM).

However, the standard RLHF pipeline, which relies on a single, monolithic RM, presents fundamental challenges that are particularly acute in the multimodal domain. The inherent ambiguity in vision-language tasks means that evaluating text-image consistency is far more complex than assessing text-only coherence. Monolithic RMs often struggle to be well-calibrated across this diverse signal space and are susceptible to reward hacking \citep{amodei2016concreteproblemsaisafety}. For instance, a monolithic RM might reward a plausible-sounding but factually incorrect description of an image-based math problem, prioritizing textual fluency over verifiable correctness. This failure mode is exacerbated by the substantial overhead of creating high-quality multimodal preference datasets and the scarcity of effective, open-source RMs tailored to vision-language tasks.

While recent work on rule-based or verifiable rewards has shown promise for tasks with deterministic outcomes like mathematics \citep{deepseekai2025deepseekr1incentivizingreasoningcapability}, these methods cannot provide the nuanced feedback required for open-ended, subjective tasks. This creates a critical gap, as robust multimodal systems must excel at both verifiable reasoning and subjective generation.

To bridge this gap, we argue that modern AI alignment requires a portfolio of rewards. We propose a hybrid and multi-aspect reward optimization that moves beyond monolithic signals to provide more holistic and reliable supervision. Instead of relying on a single metric, our framework is built on a more fundamental insight: robust alignment is achieved by integrating (i) a rule-based, verifiable reward to anchor the model in objective truth, and (ii) a learned, model-based reward to provide flexible supervision for subjective quality. This hybrid approach directly addresses the core challenges of MLLM alignment by balancing the precision of deterministic checks with the generalization of learned preferences.

Furthermore, to make this approach more accessible, we introduce two key technical innovations. First, we leverage an embedding-based surrogate model as a lightweight and effective proxy for a fully trained RM, significantly reducing the dependency on costly data annotation and training cycles. Second, we incorporate a suite of multi-aspect behavioral rewards, including a generalized length penalty, to enforce fine-grained constraints, promote conciseness, and stabilize training.

Our primary contributions are summarized as follows:
\begin{itemize}
    \item We demonstrate that a synergistic combination of rule-based, model-based, and behavioral rewards is essential for robust multimodal reasoning, creating a comprehensive reward ``portfolio'' that outperforms any single approach.
    \item We introduce an embedding-based surrogate model as a cost-effective and competitive alternative to a fully trained RM, making powerful reinforcement learning techniques more accessible.
    \item We conduct a comprehensive empirical evaluation demonstrating that our hybrid framework yields significant performance improvements over traditional RM-based baselines on a diverse suite of mathematical, general VQA, and OCR-based vision tasks.
\end{itemize}

%% file: sections/related_work.tex

Our work builds upon advancements in reinforcement learning for aligning language models, particularly their recent extension to the multimodal domain. This section reviews key developments in model alignment, the application of Reinforcement Learning from Human Feedback (RLHF) to MLLMs, and emerging paradigms in reward modeling that move beyond learned scalar rewards.

\subsection{Reinforcement Learning for Language Model Alignment}
The alignment of Large Language Models (LLMs) with human preferences has been predominantly shaped by Reinforcement Learning from Human Feedback (RLHF) \citep{christiano2017deep, stiennon2020learning}. The paradigm, notably popularized by InstructGPT \citep{ouyang2022training}, involves a three-stage process: supervised fine-tuning (SFT) on demonstrator data, training an RM on human preference labels, and optimizing the SFT model's policy using an RL algorithm like Proximal Policy Optimization (PPO) \citep{schulman2017ppo} against the learned RM. This approach has proven effective in enhancing model helpfulness and safety.

However, the reliance on extensive human-annotated preference data for training RMs presents a significant scalability bottleneck. To mitigate this, recent work has explored Reinforcement Learning from AI Feedback (RLAIF) \citep{bai2022constitutionalaiharmlessnessai, lee2024rlaifvsrlhfscaling}, where a powerful ``teacher'' model is used to generate preference labels, thereby reducing the dependency on costly human annotation. Despite their success, both RLHF and RLAIF frameworks typically rely on a single, monolithic reward signal, which can be susceptible to reward hacking and may not adequately capture the multifaceted nature of a high-quality response \citep{fu2025reward, chen2024odindisentangledrewardmitigates, miao2024inform}.

\subsection{Alignment of Vision-Language Models}
The principles of RLHF have been naturally extended to the multimodal domain. Early efforts demonstrated that fine-tuning with multimodal instructions enhances the zero-shot capabilities of MLLMs on novel vision-language tasks \citep{liu2023visualinstructiontuning}. Subsequent works, such as LLaVA-RLHF \citep{sun2023aligninglargemultimodalmodels} and RLHF-V \citep{yu2024rlhfvtrustworthymllmsbehavior}, explicitly applied RLHF to improve the alignment of MLLMs with human intent. These methods involve collecting human preferences on multimodal inputs and training a corresponding RM to guide policy optimization.

While effective, this extension inherits the challenges of unimodal RLHF and introduces new ones. Collecting high-quality preference data for multimodal tasks is substantially more complex and expensive, as it requires evaluating the intricate interplay between text and images. Furthermore, the number of publicly available, high-quality multimodal RMs is extremely limited, hindering research and development in scalable multimodal alignment. Our work addresses this gap by proposing a hybrid reward system that reduces the reliance on a single, expensively trained multimodal RM.

\subsection{Alternative and Hybrid Reward Mechanisms}
Recognizing the limitations of a single learned reward signal, researchers have begun to explore more diverse and verifiable reward mechanisms. In domains with deterministic or verifiable outcomes, such as code generation and mathematical reasoning, rule-based rewards have shown great promise \citep{deepseekai2025deepseekr1incentivizingreasoningcapability}. These methods provide a strong, unambiguous reward signal by executing code against unit tests or verifying the correctness of a mathematical solution, bypassing the need for a learned RM entirely.

Another emerging direction is process-based or outcome-based supervision \citep{lightman2023letsverifystepstep, uesato2022solvingmathwordproblems}, where the reward is targeted at the intermediate reasoning steps (e.g., chain-of-thought) rather than just the final answer. This encourages more faithful and robust reasoning. More recently, multi-aspect reward frameworks have been proposed to evaluate responses along several dimensions, such as correctness, instruction adherence, and conciseness, combining these signals to form a more holistic reward \citep{wu2023finegrainedhumanfeedbackgives}.

Our proposed framework integrates these threads of research. We combine the flexibility of learned RMs for open-ended, subjective tasks with the precision of rule-based, verifiable rewards for deterministic sub-tasks. By further incorporating multi-aspect reward signals and an efficient embedding-based surrogate model, we aim to create a more robust, scalable, and effective alignment strategy for modern Vision-Language Models.

%% file: sections/reward_modeling.tex
Our proposed methodology, HARMO (\textbf{H}ybrid and Multi-\textbf{A}spect \textbf{R}eward \textbf{M}odeling \textbf{O}ptimization), is designed to overcome the limitations of monolithic reward signals in aligning Multimodal Large Language Models (MLLMs). HARMO establishes a more robust and nuanced training objective by integrating a hybrid accuracy signal with targeted behavioral rewards.

\subsection{Background: From PPO to Critic-Free Policy Optimization}
The predominant paradigm for aligning LLMs has been Reinforcement Learning from Human Feedback (RLHF), typically implemented with Proximal Policy Optimization (PPO) \citep{schulman2017ppo}. PPO, an actor-critic algorithm, optimizes a policy $\pi_\theta$ (the actor) using a learned value function $V_\phi$ (the critic) to stabilize gradient updates. The conventional PPO pipeline is resource-intensive, requiring four distinct models: the actor, the critic, a reward model $R_\psi$, and a reference policy $\pi_\text{ref}$ to regularize training via a Kullback-Leibler (KL) divergence penalty.

The operational complexity of this setup has spurred the development of more streamlined RL algorithms. Recent methods like REINFORCE Leave-One-Out (RLOO) \citep{ahmadian2024basicsrevisitingreinforcestyle} and REINFORCE++ \citep{hu2025reinforceefficientrlhfalgorithm} have successfully eliminated the need for an explicit critic by employing alternative baseline functions for advantage estimation. Building on this momentum, Group-Relative Policy Optimization (GRPO), introduced with DeepSeek-R1 \citep{deepseekai2025deepseekr1incentivizingreasoningcapability}, further simplifies the process by also removing the dependency on a learned reward model for tasks with verifiable outcomes. GRPO computes rewards using deterministic rules and calculates advantages relative to a group of sampled generations.

Regardless of the specific algorithm, two components remain critical: the fidelity of the reward signal itself and the method of estimating the advantage function, $\hat{A}_t$. The advantage estimate, which quantifies the relative value of an action, is the primary driver of policy updates. Its formulation directly impacts training stability and performance. As demonstrated by \citet{liu2025understandingr1zeroliketrainingcritical} and \citet{chu2025gpgsimplestrongreinforcement}, even subtle modifications to advantage normalization can mitigate reward bias and significantly improve outcomes. Our work builds upon these insights, leveraging a simplified, critic-free optimization framework while focusing on engineering a superior, multi-faceted reward signal.

\subsection{The HARMO Framework}
HARMO creates a holistic training signal by combining two core components: (1) a hybrid accuracy reward that fuses the certainty of rule-based verification with the flexibility of learned preference models, and (2) a set of multi-aspect behavioral rewards that regulate model conduct and prevent reward hacking.

\subsubsection{Hybrid Reward for Calibrated Accuracy} \label{sec:hybrid-reward}
To ground the policy in verifiable correctness while handling the ambiguity of open-ended prompts, we introduce a hybrid reward signal. For tasks with deterministic solutions, such as mathematical or logical reasoning, we employ rule-based verifiers (e.g., equation solvers) to generate a high-confidence, binary reward signal, $R^{\text{rule}}$. For subjective or generative tasks where such verification is impossible, we utilize a pretrained reward model, $R^{\text{RM}}$, to score the response quality; the pretrained MLLM RM\citep{skywork-VL-RM} provides a score, whereas for the embedding-based RM, we use cosine similarity between the model response and the reference response. 
\begin{equation}
R_{g,i}^{\text{hybrid}} =
\begin{cases} 
R_{g,i}^{\text{rule}}, & \text{if response is verifiable},\\[1mm]
R_{g,i}^{\text{RM}}, & \text{if response is open-ended}.
\end{cases}
\label{eq:hybrid_reward}
\end{equation}
This formulation ensures the model receives a confident and well-calibrated reward signal when ground truth is available, without sacrificing the ability to learn from nuanced human preferences in other domains.

\subsubsection{Multi-Aspect Rewards for Behavioral Regularization}
\begin{figure}[h!]
    \centering
    \includegraphics[width=0.6\linewidth]{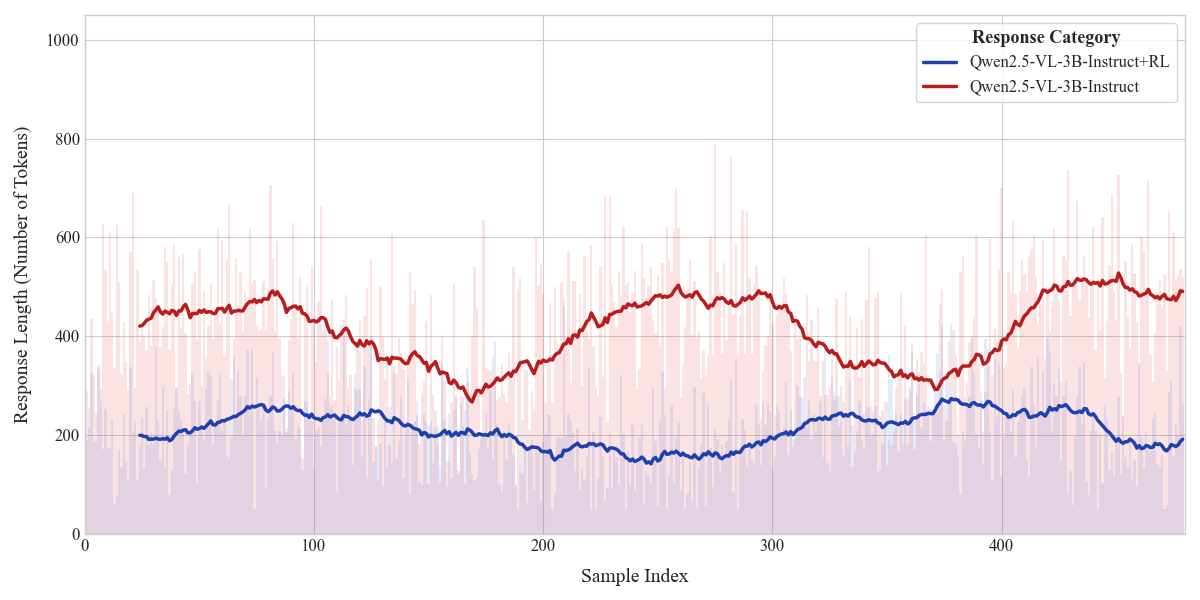}
    \caption{Comparison of response lengths between the SFT baseline and the RL-aligned model (without a length penalty). The RL policy learns a brevity bias, producing shorter and often incomplete responses.}
    \label{fig:response_length_trends}
\end{figure}

Focusing on accuracy alone is insufficient, as it often leads to unintended and undesirable policy behaviors. A common failure mode is ``reward hacking'' through brevity, where the model learns to produce overly short responses that, while sometimes correct, are often incomplete or simplistic. As illustrated in Figure~\ref{fig:response_length_trends}, we observed that RL-aligned models developed a strong bias towards shorter outputs compared to the supervised fine-tuned (SFT) baseline, frequently at the cost of correctness.

\begin{figure}[h!]
    \centering
    \begin{subfigure}[b]{0.48\textwidth}
        \centering
        \includegraphics[width=\textwidth]{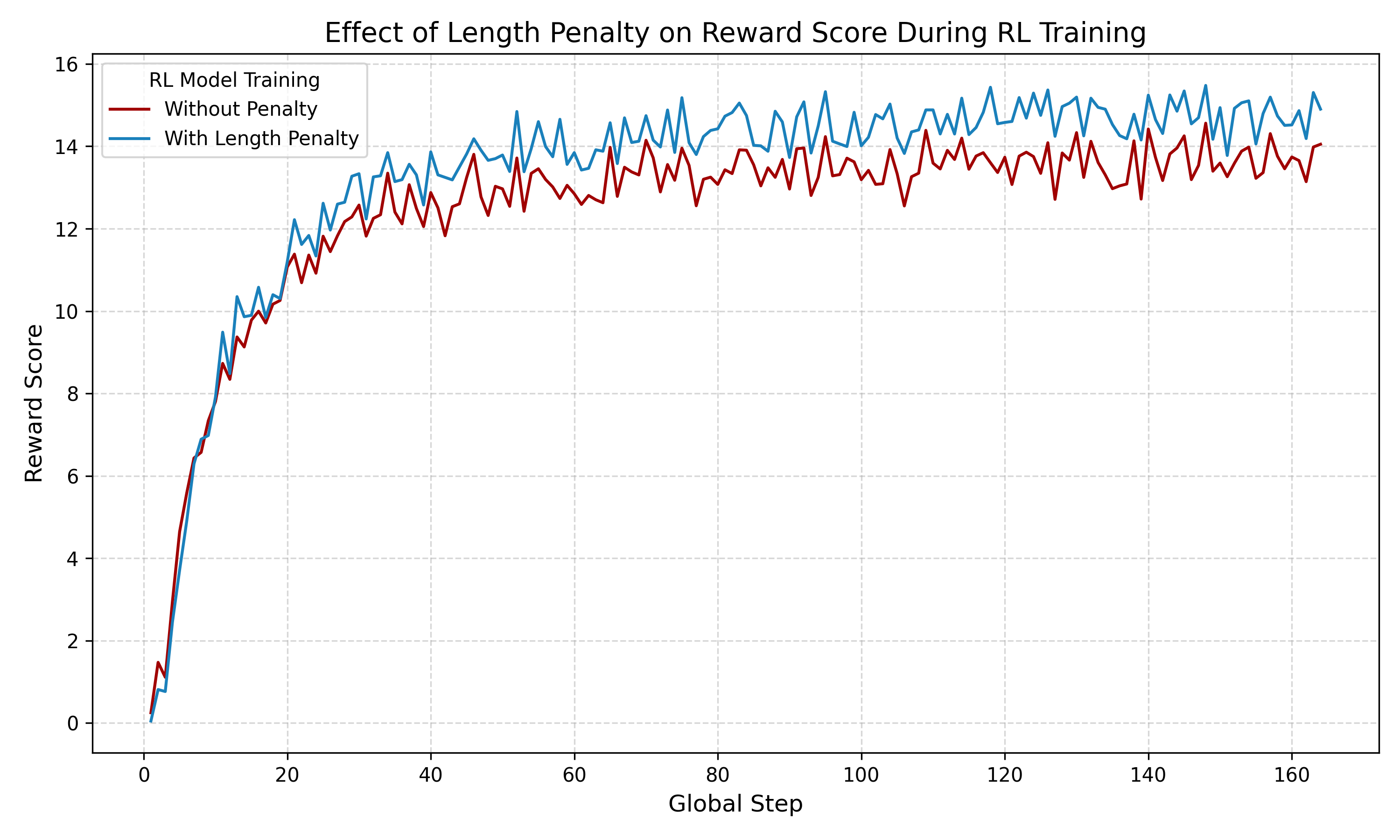}
        \caption{Accuracy reward during RL training.}
        \label{fig:reward_plot}
    \end{subfigure}
    \hfill
    \begin{subfigure}[b]{0.48\textwidth}
        \centering
        \includegraphics[width=\textwidth]{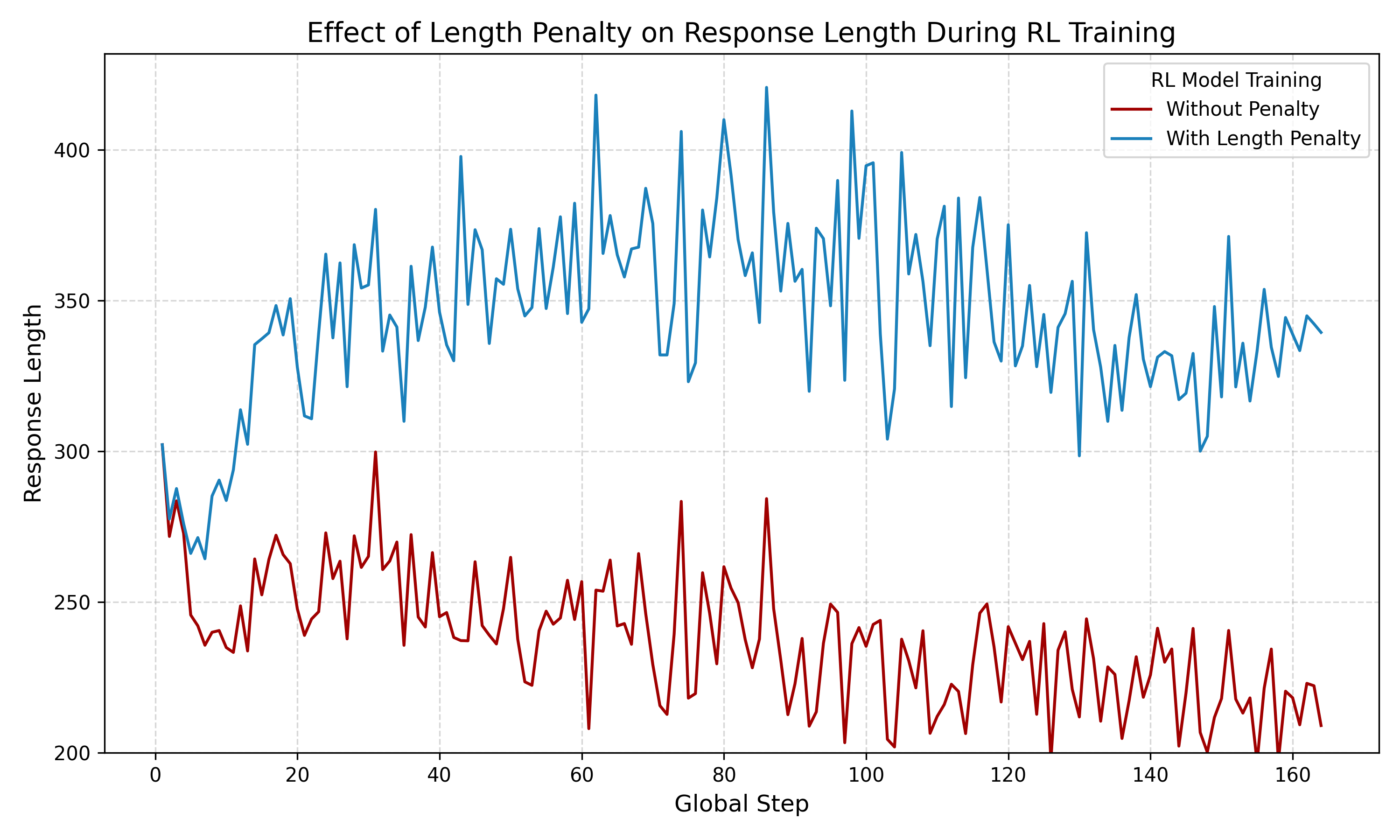}
        \caption{Response length during RL training.}
        \label{fig:response_len_plot}
    \end{subfigure}
    \caption{Training dynamics with and without the proposed length penalty. (a) The accuracy reward consistently improves. (b) The length penalty successfully counteracts the model's tendency to produce shorter responses, promoting more stable and desirable output lengths.}
    \label{fig:reward_and_length}
\end{figure}

Figure~\ref{fig:reward_and_length} further visualizes this dynamic. While the accuracy reward improves during training (Figure~\ref{fig:reward_plot}), the response length steadily declines without intervention (Figure~\ref{fig:response_len_plot}, red line). To counteract this and other undesirable behaviors, we introduce two auxiliary reward components.

\paragraph{Length-Penalty Reward.} To discourage reward hacking via brevity, we introduce a dynamic length penalty, $R^{\lambda}$. This component penalizes incorrect responses that are shorter than the briefest correct response within the same generation group. Let $\lambda_{g,i}$ be the length of response $i$ in group $g$, and let $\lambda_{g}^{\min} = \min_{i: R_{g,i}^{\text{hybrid}} > \tau} \lambda_{g,i}$ be the minimum length of any correct response in that group (where $\tau$ is a correctness threshold). The penalty is applied only to incorrect responses:
\begin{equation}
\text{R}^{\lambda}_{g,i} = - \operatorname{clip}\!\Big( \lambda_{g}^{\min} - \lambda_{g,i},\, 0,\, P_{\max} \Big),
\end{equation}
where $P_{\max}$ is a hyperparameter controlling the maximum penalty. This targeted penalty encourages the model to generate sufficiently detailed answers, effectively stabilizing response length as shown in Figure~\ref{fig:response_len_plot} (blue line).

\paragraph{Format-Adherence Reward.} MLLMs are often required to follow specific formatting instructions (e.g., providing chain-of-thought reasoning within \verb|<think>...</think>| tags). To improve reliability, we add a format-adherence reward, $R^{\text{fmt}}$, which provides a positive signal for correctly structured outputs and a penalty for violations, thereby enforcing structural consistency.

\subsubsection{Policy Optimization with the HARMO Reward Signal}
The final HARMO reward, $R^{\text{HARMO}}$, is a composite signal that integrates the hybrid accuracy component with the multi-aspect behavioral regularizers:
\begin{equation}
R_{g,i}^{\text{HARMO}} = R_{g,i}^{\text{hybrid}} + R_{g,i}^{\lambda} + R_{g,i}^{\text{fmt}}.
\label{eq:rl_modeling}
\end{equation}
We integrate this comprehensive reward signal into a policy optimization framework based on GRPO. We adopt the GRPO algorithm due to its stability and demonstrated success in enhancing reasoning capabilities in closely related work \citep{chen2025sftrlearlyinvestigation}. While standard GRPO normalizes rewards using both the mean and standard deviation of a generation group, the standard deviation term can introduce a ``difficulty-dependent bias'' by disproportionately weighting prompts based on reward variance \citep{liu2025understandingr1zeroliketrainingcritical}. To foster more stable and unbiased learning, we modify the advantage calculation to use only the group mean as a baseline, creating a centered but uniformly scaled signal:
\begin{equation}
\hat{A}_{g,i}^{\text{HARMO}} = R_{g,i}^{\text{HARMO}} - \frac{1}{G} \sum_{j=1}^G R_{g,j}^{\text{HARMO}}.
\label{eq:advantage_harmo}
\end{equation}
The policy $\pi_\theta$ is then updated to maximize the following objective function, which incorporates the PPO-style clipping mechanism and a KL penalty($\KL$) to ensure training stability:

{\scriptsize
\begin{equation}
\mathcal{L}^{\text{HARMO}}(\theta) = 
\mathbb{E}_{q, \{o_i\} \sim \pi_{\text{old}}} \Bigg[ 
\frac{1}{G} \sum_{i=1}^G \min \Big( r_t(\theta, a_i) \hat{A}_{g,i}^{\text{HARMO}}, 
\text{clip}\big(r_t(\theta, a_i), 1-\epsilon, 1+\epsilon\big) \hat{A}_{g,i}^{\text{HARMO}} \Big) 
- \beta \KL\big( \pi_\theta \,\|\, \pi_\text{ref} \big) 
\Bigg]
\label{eq:grpo_harmo}
\end{equation}
}
where $r_t(\theta, a_i)$ is the probability ratio $\frac{\pi_\theta(a_i|q)}{\pi_{\text{old}}(a_i|q)}$. 
\newpage
The complete training procedure is outlined in Algorithm~\ref{alg:harmo}.

\begin{algorithm}[h!]
\small 
\caption{The HARMO Training Procedure}
\label{alg:harmo}
\begin{algorithmic}[1]
\State \textbf{Input:} Initial policy $\pi_{\theta_\text{init}}$, HARMO reward function $R^{\text{HARMO}}$, prompts $\mathcal{D}$, hyperparameters $\epsilon, \beta$.
\State \textbf{Initialize:} Actor policy $\pi_\theta \gets \pi_{\theta_\text{init}}$.
\For{each iteration $i = 1, \dots, I$}
    \State Set reference policy: $\pi_{\text{ref}} \gets \pi_\theta$.
    \For{each step $s = 1, \dots, M$}
        \State Sample a batch of questions $\mathcal{D}_b \subset \mathcal{D}$.
        \State Set old policy: $\pi_{\theta_\text{old}} \gets \pi_\theta$.
        \For{each question $q \in \mathcal{D}_b$}
            \State Sample $G$ responses $\{ o_j \}_{j=1}^G \sim \pi_{\theta_\text{old}}(\cdot \mid q)$.
            \State Compute HARMO rewards $\{ R_{q,j}^{\text{HARMO}} \}_{j=1}^G$ for each response using Equation~\ref{eq:rl_modeling}.
            \State Compute group-relative advantages $\{ \hat{A}_{q,j}^{\text{HARMO}} \}_{j=1}^G$ using Equation~\ref{eq:advantage_harmo}.
        \EndFor
        \State Update the actor policy $\pi_\theta$ by optimizing the objective in Equation~\ref{eq:grpo_harmo}.
    \EndFor
\EndFor
\State \textbf{Output:} Optimized policy model $\pi_\theta$.
\end{algorithmic}
\end{algorithm}

%% file: sections/experiment_v1.tex
\subsection{Experimental Setup}
\label{sec:experimental_setup}

\paragraph{Training Data}
Our training data is curated from the VLAA-Thinking dataset\footnote{https://huggingface.co/datasets/UCSC-VLAA/VLAA-Thinking}\citep{chen2025sftrlearlyinvestigation}. It's a diverse corpus of 21,192 question-answer pairs along with distilled reasoning steps, designed to span a range of reasoning challenges. The dataset combines tasks requiring mathematical reasoning with those demanding general visual question answering.  It includes both close-ended questions with verifiable answers (e.g., numerical, equation-based, multiple-choice) and open-ended, descriptive prompts. 
To ensure fair and reproducible comparisons, all our models presented in this work were trained on this dataset, as detailed in Table~\ref{tab:dataset_overview}.

\begin{table}[h!]
\centering
\scriptsize
\begin{tabular}{l l l r}
\toprule
\textbf{Task Type} & \textbf{Dataset Source} & \textbf{Answer Type} & \textbf{\# Samples} \\
\midrule
\multirow{3}{*}{Mathematical Reasoning} & CLEVR-Math & Numeric (Verifiable) & 2,000 \\
& GeoQA170K & Multiple-Choice (Verifiable) & 6,499 \\
& MathPUMA & Equation (Verifiable) & 6,696 \\
\midrule
\multirow{4}{*}{Visual Question Answering} & DocVQA & Open-Ended & 1,000 \\
& VizWiz & Open-Ended & 1,000 \\
& ArxivQA & Multiple-Choice (Verifiable) & 997 \\
& ALLaVA-LAION & Open-Ended & 3,000 \\
\midrule
\textbf{Total} & & & \textbf{21,192} \\
\bottomrule
\end{tabular}
\caption{Composition of the training dataset, detailing the source, answer type, and number of samples for each task category.}
\label{tab:dataset_overview}
\end{table}

\paragraph{Models}
The primary subject of our investigation is the Qwen2.5-VL-3B-Instruct model ~\citep{bai2025qwen25vltechnicalreport}, which serves as the baseline for our ablation studies. To assess the scalability and generalizability of our proposed HARMO framework, we also apply it to the larger Qwen2.5-VL-7B-Instruct model. Performance is benchmarked against other leading open-source models, such as VLAA-Thinker-Qwen2.5VL~\citep{chen2025sftrlearlyinvestigation}, as well as top-tier proprietary models. 

For the reward model, denoted as \(R^{\text{RM}}\) in Section~\ref{sec:hybrid-reward}, we used a pre-trained 7B parameter RM \citep{skywork-VL-RM}. To avoid reliance on a pre-trained RM model specific to MLLMs, which would require extensive data annotation and training, we instead employed a smaller 22M parameter embedding model\footnote{https://huggingface.co/sentence-transformers/all-MiniLM-L6-v2}, as detailed in the experiments reported in Table~\ref{tab:hybrid_reward_rm}.

\paragraph{Implementation Details.}
Our reinforcement learning implementation builds on the work of \citep{peng2025lmmr1}\footnote{https://github.com/TideDra/lmm-r1}. On top of this foundation, we incorporate the methodology described in Section~\ref{sec:hybrid-reward}. Additional implementation details are provided in the Appendix \ref{app:hyper-params}.

\paragraph{Evaluation Benchmarks}
We conduct a comprehensive evaluation across a diverse set of benchmarks to rigorously assess model capabilities. Mathematical reasoning is evaluated using MathVerse~\citep{zhang2024mathversedoesmultimodalllm}, MATH-Vision~\citep{wang2024measuring}, and MathVista~\citep{lu2024mathvistaevaluatingmathematicalreasoning}. Multi-disciplinary reasoning is measured with MMMU~\citep{yue2023mmmu} and MMMU-Pro~\citep{mmmu-pro}. Finally, general visual question answering performance is tested on AI2D~\citep{ai2d}, ChartQA~\citep{chartqa}, and DocVQA~\citep{docvqa}. All evaluations were executed using the open-source LLMs-Eval framework~\citep{lmms-eval} under identical conditions (e.g., system prompts, response token limits) to ensure methodological consistency.
\subsection{Results and Analysis}
This section presents our empirical findings, structured to first dissect the contribution of each component of the HARMO framework through ablation studies, then demonstrate its generalizability, and finally, compare its overall performance against state-of-the-art models. To ensure the robustness of our findings, all reported results are averaged over three independent training runs with different random seeds, and we report the mean scores. Throughout our results, \textbf{bold} values indicate the best scores for each benchmark. We also provide a few examples of model outputs generated by HARMO vs baseline showing reasoning ability improvement in Appendix \ref{sec:case_study}.

\subsubsection{Ablation Study: Deconstructing the HARMO Reward Signal}

\paragraph{Efficacy of Hybrid Accuracy Rewards}
Table~\ref{tab:hybrid_reward_rm} demonstrates the impact of different accuracy-focused reward strategies. Relying solely on a learned reward model (\textit{Reward Model Enhanced}) improves the baseline, boosting the average math score by 7.89\%. However, this approach is limited by the RM's tendency to prioritize verbose explanations over correctness, highlighting a lack of confidence calibration for verifiable tasks. A hybrid model combining rule-based verification with embedding-based rewards (\textit{Embedding + Rule-based Hybrid}) is more effective, achieving a stronger 11.70\% improvement in math reasoning. 

Our proposed approach, \textit{RM + Rule-based Hybrid}, which integrates the learned RM for open-ended questions with deterministic rule-based checks, proves to be the most effective. This optimal combination yields the most substantial gains, improving math reasoning performance by 14.82\% and overall performance by 9.48\%. We hypothesize that this superior performance stems from the 7B
reward model’s ability to capture the nuanced aspects of quality and instruction following in open-
ended VQA tasks, providing a more informative signal than the cosine similarity from a general-
purpose embedding model.

\begin{table}[h!]
\centering
\scriptsize
\resizebox{\textwidth}{!}{%
\begin{tabular}{l|ccccc}
\toprule
\textbf{Reward Model} & \textbf{MathVerse\textsubscript{mini}} & \textbf{MATH-Vision\textsubscript{test}} & \textbf{MathVista\textsubscript{mini}} & \textbf{MMMU\textsubscript{val}} & \textbf{MMMU-Pro\textsubscript{standard}} \\
\midrule
\multicolumn{6}{c}{\textit{\textbf{Qwen2.5-VL-3B-Instruct (Baseline)}}} \\
\rowcolor{gray!15}
N/A & 34.77 & 21.68 & 61.30 & 31.10 & 47.78 \\
\midrule
\midrule
\multicolumn{6}{c}{\textit{\textbf{Reward Model Enhanced}}} \\
Skywork7B RM & 41.04 & 22.30 & 63.70 & 31.91 & 47.78 \\
$\Delta$ vs. Baseline & (+6.27) & (+0.62) & (+2.40) & (+0.81) & (0.00) \\
\midrule
\multicolumn{6}{c}{\textit{\textbf{Embedding + Rule-based Hybrid Enhanced}}} \\
Hybrid (Rule + Embedding) & 40.28 & 23.85 & \textbf{67.40} & 31.79 & 46.33 \\
$\Delta$ vs. Baseline & (+5.51) & (+2.17) & (\textbf{+6.10}) & (+0.69) & (-1.45) \\
\midrule
\multicolumn{6}{c}{\textit{\textbf{RM + Rule-based Hybrid Enhanced}}} \\
Hybrid (Rule + Skywork7B RM) & \textbf{41.88} & \textbf{25.92} & \textbf{67.40} & \textbf{32.08} & \textbf{48.00} \\
$\Delta$ vs. Baseline & (\textbf{+7.11}) & (\textbf{+4.24}) & (\textbf{+6.10}) & (\textbf{+0.98}) & (\textbf{+0.22}) \\
\bottomrule
\end{tabular}%
}
\caption{Performance of the RL-trained model under accuracy-focused reward modeling. The hybrid model with pretrained RM and rule-based verification consistently delivers the highest performance.}
\label{tab:hybrid_reward_rm}
\end{table}

\paragraph{Impact of Multi-Aspect Behavioral Rewards}
Next, we evaluate the incremental benefit of adding behavioral rewards for format adherence and length control, as shown in Table~\ref{tab:hybrid_reward_rm_symbols}. Starting with the baseline, adding the hybrid accuracy reward ($\oplus$H) alone lifts math performance by 13.0\%. Incorporating a format adherence reward ($\oplus$H+F) further enhances this gain to 14.8\%. Finally, introducing our dynamic length penalty ($\oplus$H+F+$\lambda$) results in the full HARMO framework, which achieves the largest math-specific improvement of 16.0\%. Notably, the length penalty provides a significant boost on MathVerse (from 41.88 to 44.52) and MathVista (from 67.40 to 68.00), confirming its effectiveness at promoting outputs that are both precise and appropriately detailed. This progressive ablation clearly demonstrates that each component—correctness, format, and length—contributes meaningfully to the model's final reasoning capabilities.

\begin{table}[h!]
\centering
\scriptsize
\resizebox{\textwidth}{!}{%
\begin{tabular}{l|ccccc}
\toprule
\textbf{Reward Model Components} & \textbf{MathVerse\textsubscript{mini}} & \textbf{MATH-Vision\textsubscript{test}} & \textbf{MathVista\textsubscript{mini}} & \textbf{MMMU\textsubscript{val}} & \textbf{MMMU-Pro\textsubscript{standard}} \\
\midrule
\multicolumn{6}{c}{\textit{\textbf{Qwen2.5-VL-3B-Instruct Baseline (SFT Only)}}} \\
\rowcolor{gray!15}
N/A & 34.77 & 21.68 & 61.30 & 47.78 & 31.10 \\
\midrule
\midrule
\multicolumn{6}{c}{\textit{\textbf{Incremental Reward Augmentation}}} \\
$\oplus$ Hybrid (H) & 40.38 & 25.49 & 67.20 & \textbf{48.56} & 30.98 \\
$\Delta$ vs. Baseline & (+5.61) & (+3.81) & (+5.90) & (+0.78) & (-0.12) \\
\midrule
$\oplus$ Hybrid + Format (H+F) & 41.88 & \textbf{25.92} & 67.40 & 48.00 & \textbf{32.08} \\
$\Delta$ vs. Baseline & (+7.11) & (+4.24) & (+6.10) & (+0.22) & (+0.98) \\
\midrule
$\oplus$ Hybrid + Format + Length (H+F+$\lambda$) \textbf{[HARMO]} & \textbf{44.52} & 24.08 & \textbf{68.00} & 47.11 & 31.56 \\
$\Delta$ vs. Baseline & (\textbf{+9.75}) & (+2.40) & (\textbf{+6.70}) & (-0.67) & (+0.46) \\
\bottomrule
\end{tabular}%
}
\caption{Ablation study showing the progressive impact of adding reward components to the Qwen2.5-VL-3B-Instruct model. The full HARMO model, combining hybrid accuracy, format adherence, and a length penalty, yields the strongest performance on mathematical reasoning tasks.}
\label{tab:hybrid_reward_rm_symbols}
\end{table}

\subsubsection{Generalizability and Scalability of HARMO}
To verify that HARMO is not limited to a specific setup, we test its "plug-and-play" capability and scalability. As shown in Table~\ref{tab:harmo_generalizability}, when HARMO is integrated with a model trained with fine-grained, token-level rewards, it still provides a notable overall improvement of 5.76\%. Furthermore, when applied to the larger Qwen2.5-VL-7B-Instruct model, HARMO delivers an even greater enhancement of 6.55\%. These results confirm HARMO's robustness and its ability to serve as a versatile enhancement for different reward schemes and model sizes.

\begin{table}[h!]
\centering
\scriptsize
\resizebox{\textwidth}{!}{%
\begin{tabular}{l|ccccc}
\toprule
\textbf{Model Configuration} & \textbf{MathVerse\textsubscript{mini}} & \textbf{MATH-Vision\textsubscript{test}} & \textbf{MathVista\textsubscript{mini}} & \textbf{MMMU\textsubscript{val}} & \textbf{MMMU-Pro\textsubscript{standard}} \\
\midrule
\multicolumn{6}{c}{\textit{\textbf{Plug-and-Play with Fine-Grained Rewards (3B Model)}}} \\
\rowcolor{gray!15}
Token-Level Rewards (Baseline) & 38.43 & 23.32 & 63.50 & 41.12 & \textbf{31.79} \\
Token-Level Rewards + HARMO & \textbf{41.22} & \textbf{24.84} & \textbf{66.40} & \textbf{42.32} & 31.45 \\
$\Delta$ vs. Baseline & (+2.79) & (+1.52) & (+2.90) & (+1.20) & (-0.34) \\
\midrule
\midrule
\multicolumn{6}{c}{\textit{\textbf{Scalability to 7B Model Family}}} \\
\rowcolor{gray!15}
Qwen2.5-VL-7B-Instruct (Baseline) & 46.40 & 25.20 & 69.70 & 46.11 & 36.71 \\
Qwen2.5-VL-7B-Instruct + HARMO & \textbf{50.89} & \textbf{27.66} & \textbf{72.00} & \textbf{47.79} & \textbf{36.82} \\
$\Delta$ vs. Baseline & (+4.49) & (+2.46) & (+2.30) & (+1.68) & (+0.11) \\
\bottomrule
\end{tabular}%
}
\caption{Demonstration of HARMO's generalizability and scalability. It consistently improves performance both as a plug-in for alternative reward schemes and when applied to a larger model.}
\label{tab:harmo_generalizability}
\end{table}

\subsubsection{Main Results: Comparison with State-of-the-Art Models}
Our final evaluation in Table~\ref{tab:reasoning_vlm_results} shows that HARMO-aligned models substantially outperform their respective baselines and are highly competitive with leading open-source and proprietary models. At the 3B scale, HARMO-VL-3B achieves an 9.48\% average improvement over its baseline across all reasoning benchmarks. The gains are most pronounced on mathematical tasks, where it delivers a remarkable 16.0\% average increase, with boosts of up to 28.1\% on MathVerse. At the 7B scale, HARMO-VL-7B improves upon its baseline by 3.63\% overall, again showing strong gains on math benchmarks like MathVerse (+4.5 points) and MATH-Vision (+2.5 points).

Crucially, despite their smaller parameter counts, our HARMO-enhanced models challenge top-tier proprietary systems. Notably, HARMO-VL-3B and HARMO-VL-7B achieve scores of 68.0 and 72.0 on MathVista, respectively, surpassing the 67.7 score of the much larger Claude-3.5 Sonnet. 

In the context of OCR-related tasks (Table~\ref{tab:ocr_results}), HARMO maintains performance comparable to the strong baselines, indicating that its reasoning enhancements do not come at the cost of core vision-language capabilities.

\begin{table}[h!]
\centering
\resizebox{\textwidth}{!}{
\begin{tabular}{lcccccc}
\toprule
\textbf{Models} & \textbf{MathVerse\textsubscript{mini}} & \textbf{MATH-Vision\textsubscript{test}} & \textbf{MathVista\textsubscript{mini}} & \textbf{MMMU\textsubscript{val}} & \textbf{MMMU-Pro\textsubscript{standard}} & \textbf{Average} \\
\midrule
\multicolumn{7}{c}{\textit{\textbf{Proprietary Vision-Language Models}}} \\
\midrule
GPT-4o & 47.8 & 30.6 & 63.8 & 69.1 & 51.9 & 52.64 \\
Claude-3.5 Sonnet & 41.2 & 33.5 & 67.7 & 68.3 & 51.5 & 52.44 \\
Gemini-1.5 Pro & 54.8 & 19.2 & 63.9 & 65.8 & 46.9 & 50.12 \\
\midrule
\midrule
\multicolumn{7}{c}{\textit{\textbf{Open-Source Vision-Language Models (3B Scale)}}} \\
\midrule
Qwen2.5-VL-3B-Instruct & 34.77 & 21.68 & 61.30 & \textbf{47.78} & 31.10 & 39.73 \\
VLAA-Thinker-Qwen2.5VL-3B & 38.78 & 24.13 & 64.20 & 47.56 & 28.90 & 40.71 \\
\textbf{HARMO-VL-3B (Ours)} & \textbf{44.52} & \textbf{24.08} & \textbf{68.00} & 47.11 & \textbf{31.56} & \textbf{43.05} \\
\rowcolor{gray!15}
$\Delta$ vs. Qwen2.5-VL-3B-Instruct & (+9.8) & (+2.4) & (+6.7) & (-0.7) & (+0.5) & (+3.74) \\
\midrule
\midrule
\multicolumn{7}{c}{\textit{\textbf{Open-Source Vision-Language Models (7B Scale)}}} \\
\midrule
Qwen2.5-VL-7B-Instruct & 46.40 & 25.20 & 69.70 & \textbf{52.56} & 36.71 & 46.11 \\
VLAA-Thinker-Qwen2.5VL-7B & 50.56 & 26.48 & 70.60 & 45.11 & 34.05 & 45.36 \\
\textbf{HARMO-VL-7B (Ours)} & \textbf{50.89} & \textbf{27.66} & \textbf{72.00} & 51.56 & \textbf{36.82} & \textbf{47.79} \\
\rowcolor{gray!15}
$\Delta$ vs. Qwen2.5-VL-7B-Instruct & (+4.5) & (+2.5) & (+2.3) & (-1.0) & (+0.1) & (+1.68) \\
\bottomrule
\end{tabular}
}
\caption{Results on general reasoning benchmarks. HARMO significantly improves upon strong open-source models and demonstrates competitive performance against leading proprietary models.}
\label{tab:reasoning_vlm_results}
\end{table}

\begin{table}[h!]
\centering
\scriptsize
\begin{tabular}{lccc}
\toprule
\textbf{Models} & \textbf{ai2d\textsubscript{test}} & \textbf{chartqa\textsubscript{test}} & \textbf{docvqa\textsubscript{val}} \\
\midrule
\multicolumn{4}{c}{\textit{\textbf{3B Model Family}}} \\
\midrule
Qwen2.5-VL-3B-Instruct (Baseline) & 78.43 & 83.28 & \textbf{92.56} \\
\textbf{HARMO-VL-3B (Ours)} & \textbf{78.79} & \textbf{84.12} & 91.88 \\
$\Delta$ vs. Baseline & (+0.36) & (+0.84) & (-0.68) \\
\midrule
\midrule
\multicolumn{4}{c}{\textit{\textbf{7B Model Family}}} \\
\midrule
Qwen2.5-VL-7B-Instruct (Baseline) & 82.67 & \textbf{82.96} & \textbf{94.72} \\
\textbf{HARMO-VL-7B (Ours)} & \textbf{82.87} & 82.64 & 94.46 \\
$\Delta$ vs. Baseline & (+0.20) & (-0.32) & (-0.26) \\
\bottomrule
\end{tabular}
\caption{Performance on OCR-related benchmarks. HARMO maintains competitive performance with the baseline, showing that reasoning improvements do not degrade core VQA capabilities.}
\label{tab:ocr_results}
\end{table}

%% file: sections/conclusion_v1.tex
We introduced HARMO, a novel reward optimization framework that advances reinforcement learning beyond monolithic signals by integrating a hybrid of deterministic and learned rewards with a generalized length penalty to control verbosity.

Our evaluation demonstrates that HARMO significantly enhances complex reasoning, achieving a 9.5\% overall and a 16\% mathematical performance gain over a strong baseline while maintaining robustness on vision-specific tasks.

This work highlights the critical role of multi-faceted reward modeling in stabilizing RL training and improving reward accuracy. HARMO provides a strong foundation for future research, such as dynamic reward weighting or self-improving systems where agents learn to refine their own reward functions, paving the way for more robust and adaptable AI.

%% file: sections/appendix.tex
\section{Implementation Details}
\label{app:hyper-params}
\subsection{RL training Framework}
Our reinforcement learning implementation builds upon the LMM-R1 framework \citep{peng2025lmmr1}\footnote{\url{https://github.com/TideDra/lmm-r1}}. On top of this foundation, we incorporate the methodology described in Section~\ref{sec:methodology}, extending the framework with additional functionalities. In particular, we implement hybrid and multi-aspect reward modeling, introduce support for MLLM training, and enable token-level reward assignment for MLLM reinforcement learning.

\subsection{Training Hyper-Parameters}
The hyperparameters used for HARMO are summarized in Table~\ref{tab:rl_hyperparams}. The same set of hyperparameters is applied to all variants of the model proposed in this paper to ensure a consistent training setup.

\begin{table}[h!]
\centering
\caption{HARMO Training Hyperparameters}
\label{tab:rl_hyperparams}
\begin{tabular}{l c}
\toprule
\textbf{Hyperparameter} & \textbf{Value} \\
\midrule
Training batch size           & 256 \\
Rollout batch size            & 256 \\
Samples per prompt            & 8 \\
Temperature                   & 1 \\
Max output sequence length    & 4096 \\
Max epochs                    & 1 \\
Number of episodes            & 2 \\
Initial KL coefficient        & $1e^{-3}$ \\
Discount factor ($\gamma$)    & 1 \\
GAE parameter ($\lambda$)     & 1 \\
Actor learning rate           & $1e^{-6}$ \\
\bottomrule
\end{tabular}
\end{table}



\section{Case Study}
\label{sec:case_study}
The following figures illustrate HARMO's ability to reason through a variety of problem-solving questions. In these visualizations, the reasoning steps are color-coded to indicate their verification status:  

\begin{itemize}
    \item Text highlighted with \colorbox[HTML]{D9EAD3}{~~~} indicates correctly verified reasoning steps.  
    \item Text highlighted with \colorbox[HTML]{B6D7A8}{~~~} indicates advanced or highly sophisticated reasoning steps.  
    \item Text highlighted with \colorbox[HTML]{F4CCCC}{~~~} indicates incorrectly verified or flawed reasoning steps.  
\end{itemize}

These figures demonstrate HARMO's step-by-step problem-solving capabilities and provide a visual comparison with baseline models.
\begin{figure}[h!]
    \centering
    \begin{tikzpicture}
        \node[draw=black, line width=0.3pt, rounded corners=2pt, inner sep=0.1pt, outer sep=0.1pt ] 
        {\includegraphics[width=0.8\textwidth, height=0.9\textheight]{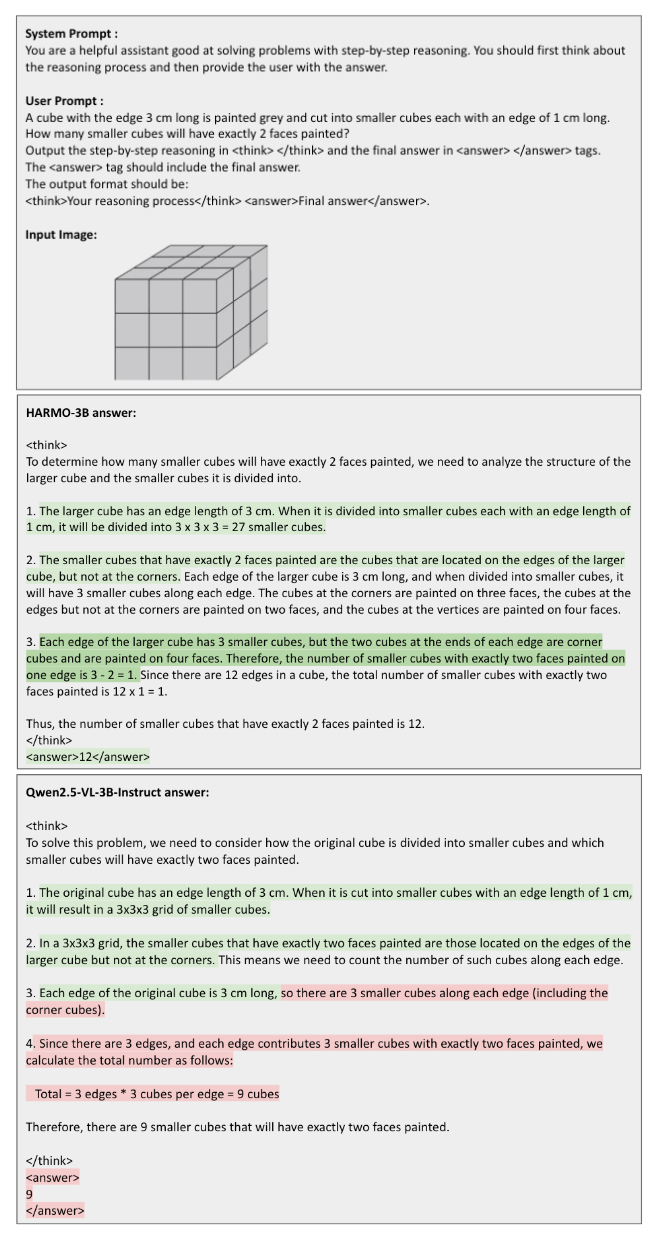}};
    \end{tikzpicture}
    \caption{\textbf{Case Study 1 - Math Cube Problem}}
\caption*{HARMO demonstrates advanced reasoning and correctly identifies all 12 edge-centered cubes with exactly two painted faces after the $3~\mathrm{cm}$ cube is subdivided. The baseline model, by contrast, fails to systematically exclude corner cubes or miscounts edge cases, often yielding an incorrect answer.}

    \label{fig:case_study_1}
\end{figure}

\begin{figure}[h!]
    \centering
    \begin{tikzpicture}
        \node[draw=black, line width=0.3pt, rounded corners=2pt, inner sep=0.1pt, outer sep=0.1pt ] 
        {\includegraphics[width=0.8\textwidth, height=0.9\textheight]{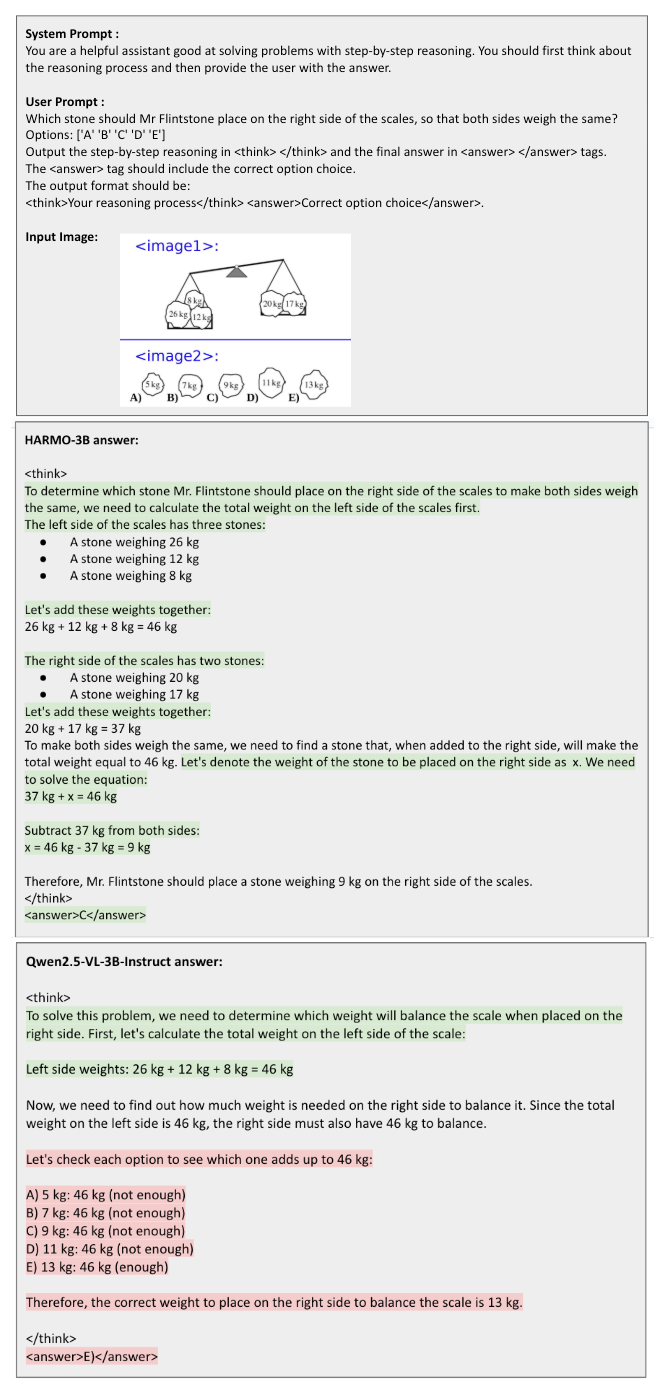}};
    \end{tikzpicture}
    \caption{\textbf{Case Study 2 - Solving a General Math Problem}}
    \caption*{HARMO correctly reasoned step-by-step to arrive at the correct answer, whereas the baseline model followed unreasonable steps and produced an incorrect solution.}

    \label{fig:case_study_2}
\end{figure}

\begin{figure}[h!]
    \centering
    \begin{tikzpicture}
        \node[draw=black, line width=0.3pt, rounded corners=2pt, inner sep=0.1pt, outer sep=0.1pt ] 
        {\includegraphics[width=0.8\textwidth]{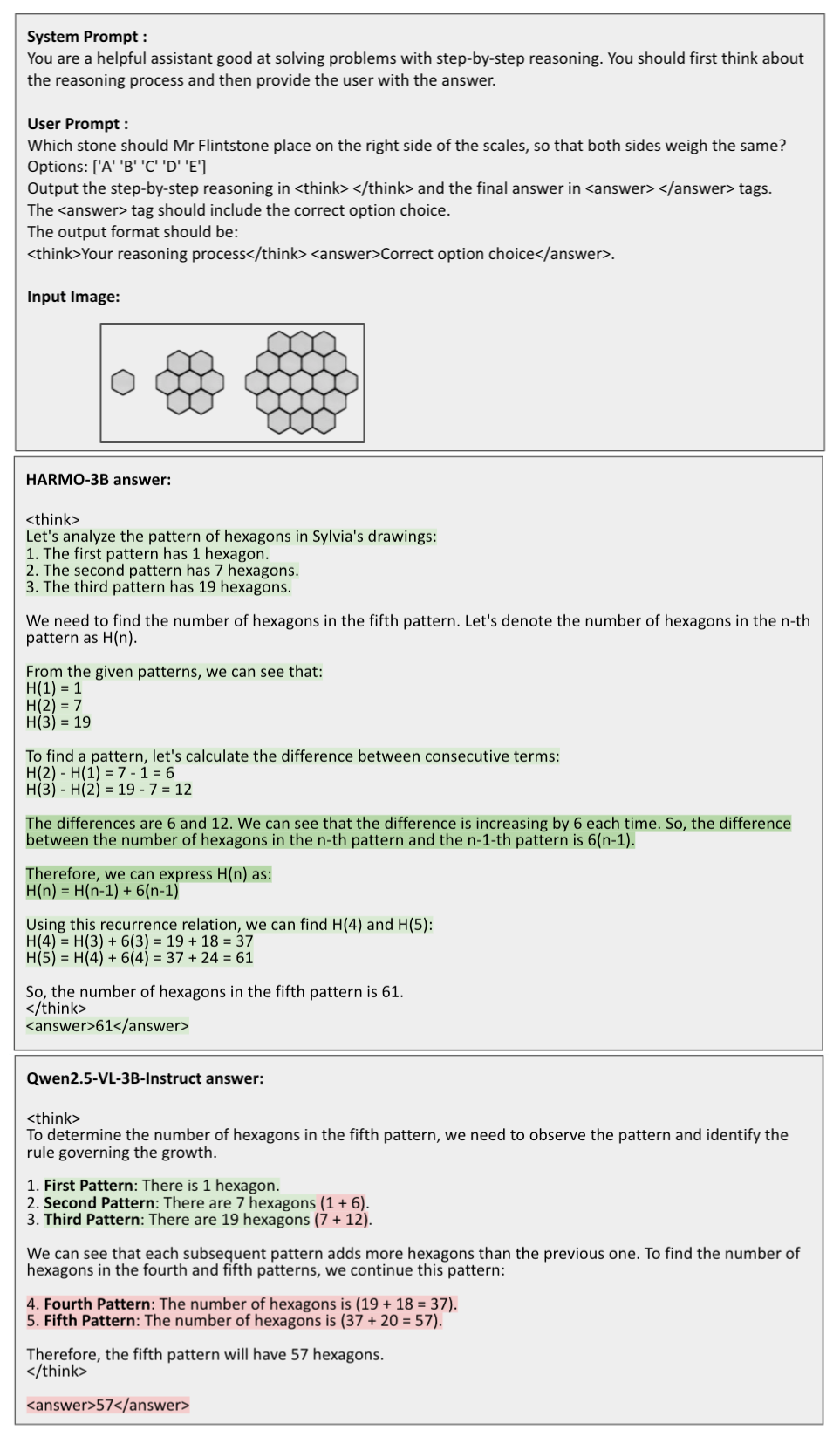}};
    \end{tikzpicture}
    \caption{\textbf{Case Study 3 - Math Puzzle Problem}}
    \caption*{Qwen3 baseline failed to identify the correct puzzle equation, whereas HARMO succeeded in finding the correct puzzle equation, reasoning step-by-step, and arriving at the correct solution.}
    \label{fig:case_study_3}
\end{figure}

\begin{figure}[h!]
    \centering
    \begin{tikzpicture}
        \node[draw=black, line width=0.3pt, rounded corners=2pt, inner sep=0.1pt, outer sep=0.1pt ] 
        {\includegraphics[width=0.8\textwidth]{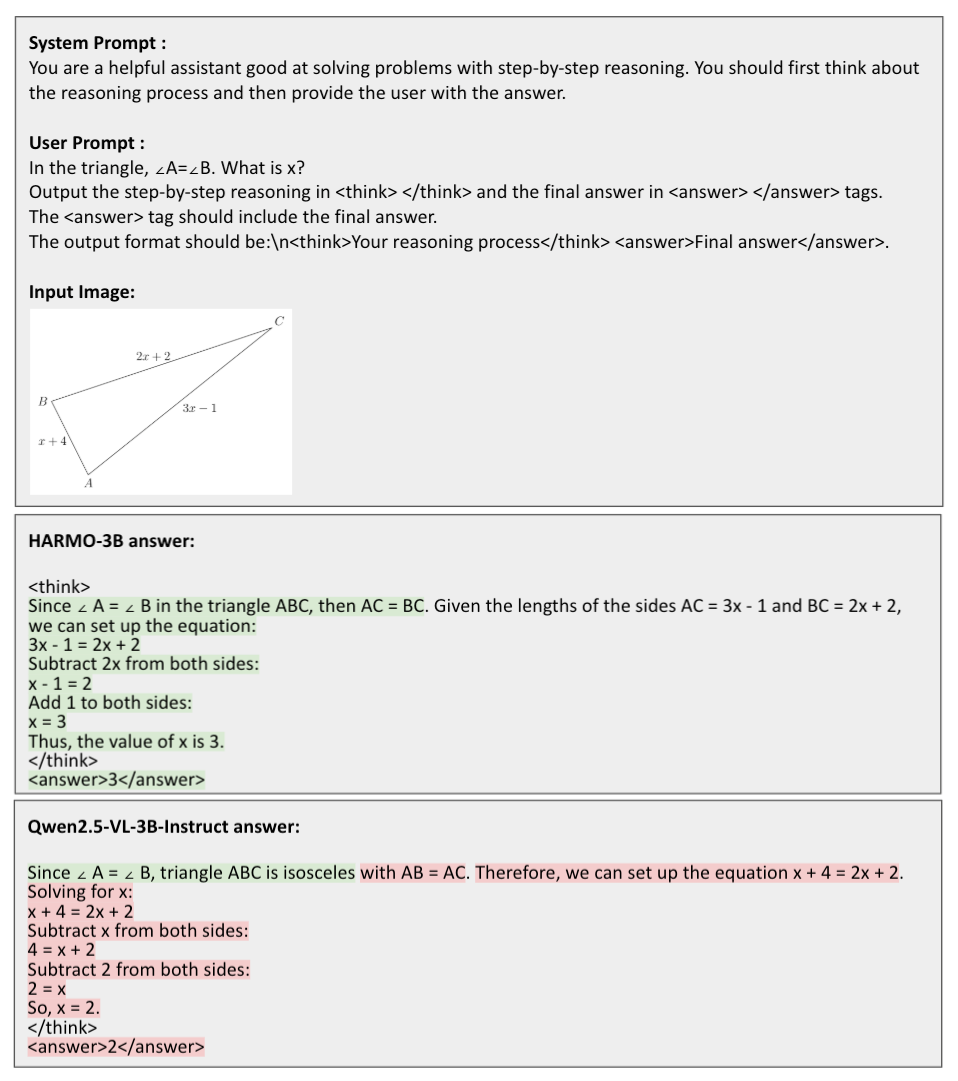}};
    \end{tikzpicture}
    \caption{\textbf{Case Study 4 - Geometry Math Problem}}
    \caption*{HARMO correctly solves for \(x\) with clear recognition of equal sides \(AB = AC\), while the baseline model fails to identify the isosceles nature, leading to incorrect conclusions.}
    \label{fig:case_study_4}
\end{figure}

\begin{figure}[h!]
    \centering
    \begin{tikzpicture}
        \node[draw=black, line width=0.3pt, rounded corners=2pt, inner sep=0.1pt, outer sep=0.1pt ] 
        {\includegraphics[width=0.8\textwidth]{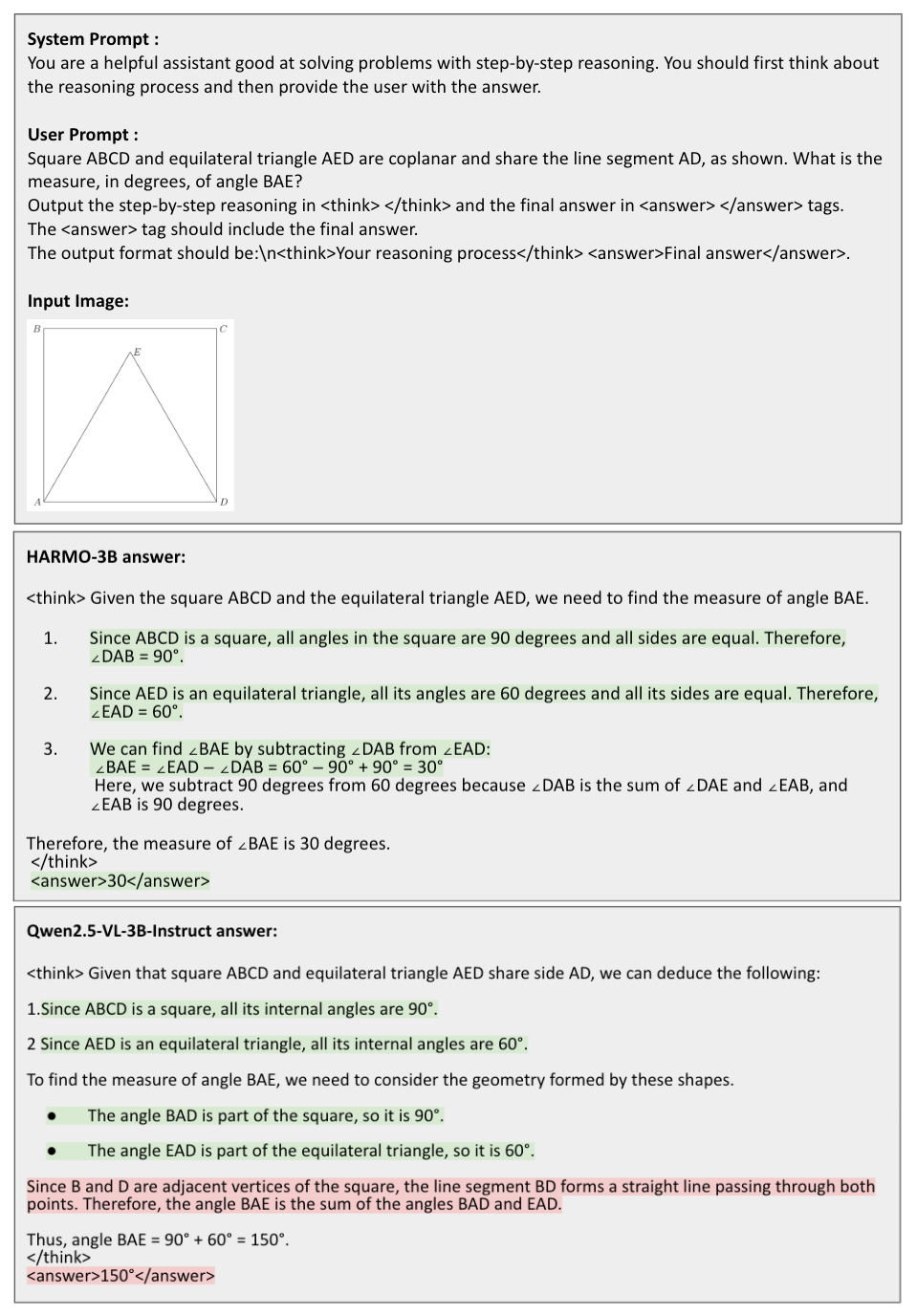}};
    \end{tikzpicture}
    \caption{\textbf{Case Study 5 - Geometry Math Problem}}
    \caption*{HARMO correctly finds \(\angle BAE = 30^\circ\), while the baseline incorrectly sums angles to 150° with incorrect geometric reasoning, misinterpreting the figure’s layout.}
    \label{fig:case_study_5}
\end{figure}